\documentclass[conf]{new-aiaa}
\usepackage[utf8]{inputenc}
\usepackage{graphicx}
\usepackage{amsfonts,amsmath}
\usepackage[version=4]{mhchem}
\usepackage{siunitx}
\usepackage{longtable,tabularx}
\usepackage{adjustbox}
\usepackage{dsfont}
\usepackage{subcaption}
\usepackage{comment}
\usepackage{svg}
\usepackage{gensymb}
\usepackage{multirow}
\usepackage{cleveref}
\graphicspath{{figs/}}
\usepackage{array}
\newcolumntype{L}[1]{>{\raggedright\let\newline\\\arraybackslash\hspace{0pt}}m{#1}}
\newcolumntype{C}[1]{>{\centering\let\newline\\\arraybackslash\hspace{0pt}}m{#1}}
\newcolumntype{R}[1]{>{\raggedleft\let\newline\\\arraybackslash\hspace{0pt}}m{#1}}

\newcommand\norm[1]{\left\lVert#1\right\rVert}

\title{Physics-Constrained Generative Artificial Intelligence for Rapid Takeoff Trajectory Design}

\author{Samuel Sisk\footnote{Doctoral Student, Department of Mechanical and Aerospace Engineering.}}
\author{Xiaosong Du\footnote{Corresponding Author, Assistant Professor, Department of Mechanical and Aerospace Engineering, AIAA Member.}}
\affil{Missouri University of Science and Technology, Rolla, MO, 65409, USA}

\begin{document}

\maketitle


\begin{abstract}
The urban air mobility (UAM) industry is rapidly growing to alleviate regional transportation congestion.
Electric vertical takeoff and landing (eVTOL) aircraft plays a critical role in this growth due to their efficiency and reduced operating cost.
However, excessive energy demands by the takeoff phase impact the practicality of the aircraft.
Conventional multidisciplinary analysis and optimization (MDAO) identifies a minimum energy trajectory but can be computationally intensive due to iteratively evaluating high-fidelity simulation models.
In addition, complex constraints in practical MDAO pose crucial challenges to the already demanding process.
Surrogate models enable efficient design optimization, but complex constraints could prohibit surrogate-based MDAO from finding the optimal design. 
This work proposes a physically constrained generative artificial intelligence model, \emph{i.e.}, physics-constrained generative adversarial networks (physicsGAN), to intelligently parameterize the takeoff control profiles of an eVTOL aircraft and to transform the original design space to a feasible space.
Specifically, the transformed feasible space refers to a space where all designs directly satisfy all design constraints.
The physicsGAN-enabled surrogate-based takeoff trajectory design framework was demonstrated on the Airbus $A^3$ Vahana.
The physicsGAN generated only feasible control profiles of power and wing angle in the feasible space with around 98.9\% of designs satisfying all constraints.
The proposed design framework obtained 99.6\% accuracy compared with simulation-based optimal design and took only 2.2 seconds, which reduced the computational time by around 200 times.
Meanwhile, data-driven GAN-enabled surrogate-based optimization took 21.9 seconds using a derivative-free optimizer, which was around an order of magnitude slower than the proposed framework. 
Moreover, the data-driven GAN-based optimization using gradient-based optimizers could not consistently find the optimal design during random trials and got stuck in an infeasible region, which is problematic in real practice.
Therefore, the proposed physicsGAN-based design framework outperformed data-driven GAN-based design to the extent of efficiency (2.2 seconds), optimality (99.6\% accurate), and feasibility (100\% feasible).
According to the literature review, this is the first physics-constrained generative artificial intelligence enabled by surrogate models.
\end{abstract}

\section{Introduction}
\label{sec:introduction}
Urban air mobility (UAM) is expected to transport over 80,000 people per year while experiencing rapid growth in the near term, with the market being worth more than two billion dollars~\cite{goyal2018urban}.
\citet{vascik2017systems} analyzed existing UAM companies and discovered that they primarily utilized helicopters, which resulted in a large cost per passenger and reduced commuter market share.
Meanwhile, \citet{duffy2017study} reviewed the benefits of electric vertical takeoff and landing (eVTOL) aircraft and noticed that they possessed reduced operating cost compared to existing helicopters.
Industry companies are producing eVTOL aircraft to fill this demand, such as Ehang, Joby, Aurora, Lilium, and Airbus, to name a few~\cite{goyal2018urban}.

The primary concerns for eVTOL are the endurance and range due to battery density, the thermal constraints of the battery and motors, and the noise pollution~\cite{yang2021challenges,sripad2021promise,chauhan2020tilt,edwards2020evtol,harrison2019using}.
These problems can be alleviated using tools, such as multidisciplinary analysis and optimization (MDAO), which balances the needs of multiple subsystems while identifying an optimal takeoff design~\cite{chauhan2020tilt,kaneko2023simultaneous}.
However, these optimizations can be high-dimensional, highly constrained, and reliant on high-fidelity simulations, which makes MDAO computationally expensive or even impossible~\cite{koziel2011surrogate}.

Conventional optimization approaches tend to convert the original problems to a series of simplified subproblem that are more straightforward to solve.
Penalty methods convert constrained optimization to unconstrained optimization using penalty functions which are straightforward for implementation but suffer from maintaining feasibility~\cite{mdobook}.
Interior point methods and sequential quadratic programming are considered state-of-the-art but still need to repeatedly deal with simplified constraints in a series of subproblems~\cite{Karmarkar1984,Gill2002a}. 

Artificial intelligence (AI) methods, including generative AI and surrogate models, have enabled multiple new optimization architectures~\cite{mlaso,Peherstorfer2018a}.  
Generative AI, especially GAN series models, permits intelligent parameterization for engineering design~\cite{du2021rapid}. 
\citet{chen2018_bezier} introduced GAN into for airfoil parameterization and enabled surrogate-based design optimization on lower-dimensional design space. 
\citet{yeh2024transfer} developed a novel surrogate model using transfer learning-based generative adversarial networks (GAN) for regression tasks and managed to directly predict optimal takeoff trajectories of electric drones based on design requirements.

Recent work has added both physics and constraints to generative AI methods to ensure generated data meets real-world requirements.
\citet{dan2020generative} used a GAN model to determine novel chemical compounds, which generated samples with 80-90\% feasibility.
\citet{yang2020physics} introduced physics-informed GAN, which they used to solve stochastic differential equations in both forward and inverse problems with up to 120 dimensions.
Similarly, \citet{yu2024physics} added a physics loss function to GAN, which allowed them to identify the probabilistic distribution of structural parameters for real-world applications.
On the other hand, \citet{engel2017latent} developed a method to generate latent constraints for an unconditionally trained GAN, which allowed them to generate outputs with desirable features without retraining the model.
Additionally, \citet{di2020efficient} developed constrained adversarial networks, which incorporated constraints into the training process of GAN.
They showed that the developed model produced feasible results more reliably than a vanilla GAN.
However, these prior work cannot effectively incorporate physical constraints due to the excessive cost of computing constraints during training. 

This work proposes a physics-constrained GAN model, termed physicsGAN, enabled by surrogate models to intelligently generate only feasible takeoff trajectories for eVTOL aircraft.
The physicsGAN will be trained on a feasible dataset provided by data-driven GAN and surrogate models.
Specifically, physicsGAN will be penalized during the training if generated data violates surrogate-predicted constraints. 
A successful training on physicsGAN is expected to transform the original design requirement space to a feasible space, where all constraints are satisfied and unconstrained optimization can be directly conducted.
Please refer to Sec.~\ref{sec:ConGAN} for more details. 

The rest of this paper is organized as follows.
Section \ref{sec:methodology} describes the simulation models, surrogate models, a data-driven twin-generator GAN (twinGAN), and the physicsGAN model used for this work.
Section \ref{sec:results} demonstrates the proposed approach on takeoff trajectory design of an eVTOL drone and compares the performance against simulation-based optimization and twinGAN-enabled surrogate-based optimization. 
Section \ref{sec:conclusion} concludes this paper with the main contributions and future work.

\section{Methodology}
\label{sec:methodology}
\begin{figure}[htbp!]
    \centering
    \adjustbox{trim=0.5cm 5.4cm 2cm 4.1cm,clip=true}{ 
    \includegraphics[scale=0.60]{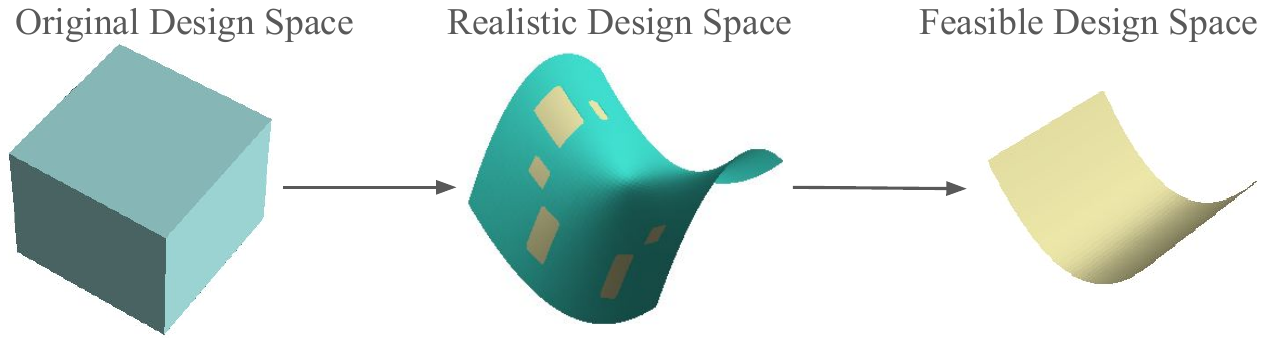}}
    \caption{The original design requirement space (left) is represented by the cube. 
    The cube's dimension is reduced to a realistic design requirement space (middle) by generating only realistic shapes, which can be achieved using the twinGAN model. The realistic space contains irregular feasible (tan) and infeasible (blue) regions. 
    The physicsGAN delivers the feasible design requirement space (right) where all constraints are satisfied.} 
    \label{fig:DesignSpace}
\end{figure}

Section \ref{sec:Sim Model} introduces the simulation models, which were originally introduced in~\cite{chauhan2020tilt,sisk2024surrogate}, used to propagate the eVTOL trajectory and generate the reference and training data for the GAN and surrogate models.
Section \ref{sec:TwinGAN} presents the twinGAN model~\cite{sisk2024surrogate}, which is used to generate realistic takeoff trajectories.
Section \ref{sec:Surrogates} describes the surrogate models used to predict the states of the aircraft during the takeoff process.
Section \ref{sec:ConGAN} elaborates on the physicsGAN model.
Figure~\ref{fig:DesignSpace} illustrates the desired transformation of the design requirement space using a simple example.

\subsection{Drone and Simulation Model}
\label{sec:Sim Model}
\begin{figure}[htbp!]
    \centering
    \label{fig:simulation}
    \begin{subfigure}{.5\textwidth}
        \centering
        \includegraphics[scale=0.6]{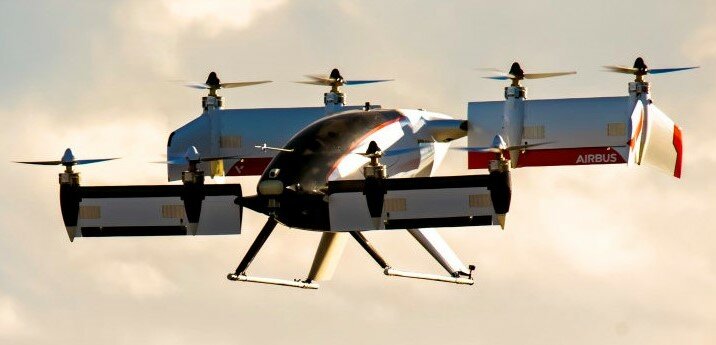}
        \caption{}
        \label{fig:Vahana}
    \end{subfigure}%
    \begin{subfigure}{.5\textwidth}
        \centering
        \includegraphics[scale=0.6]{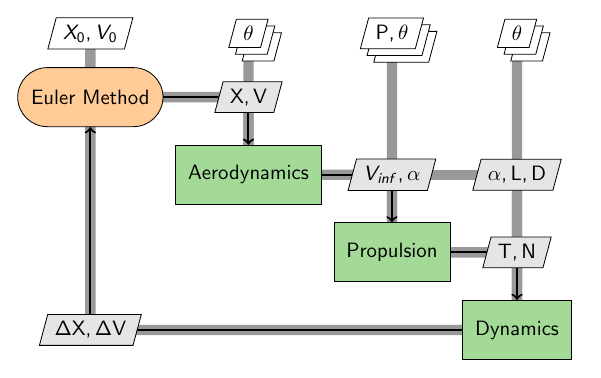}
        \caption{}
        \label{fig:sim_model}
    \end{subfigure}%
    \caption{The eVTOL model used in this work: (a) the Airbus $A^3$ Vahana eVTOL drone; (b) the model process diagram shows the flow of information through the simulation model.
    The simulation model (b) propagates the position ($X$) and velocity ($V$) by calculating the change for each time step.
    To do so, it calculates the lift and drag ($L$ and $D$), using the aerodynamics subsystem where $\alpha$ is the angle of attack, and the thrust and normal force ($T$ and $N$), using the propulsion subsystem, on the aircraft.
    The dynamics subsystem calculates the acceleration due to the forces acting on the aircraft and returns the change in position and velocity.}
\end{figure}

The Airbus $A^3$ Vahana (Fig.~\ref{fig:Vahana}) is a tandem, rotating wing eVTOL with distributed electric propulsion, which serves as the basis of the simulation model (Fig.~\ref{fig:sim_model}).
The simulation model was originally developed by \citet{chauhan2020tilt} and used in previous work~\cite{sisk2024surrogate,yeh2024transfer}.
It is derived from first-principles and augmented by empirical models.
The empirical models enable it to predict the lift and drag of the stalled wings and the normal force from the propellers.
The model states are propagated using Euler's method.
The propagated trajectory is optimized by varying the wing angle and power supplied to the motors to minimize the total energy consumed during takeoff.
The trajectory must adhere to constraints on the final displacement and velocity.
Additionally, path constraints are set on the minimum vertical displacement and maximum acceleration for passengers' comfort and safety (Sec.~\ref{sec:results}).

\subsection{Twin-Generator Generative Adversarial Network}
\label{sec:TwinGAN}
\begin{figure}
    \centering
    \includegraphics[scale=0.75]{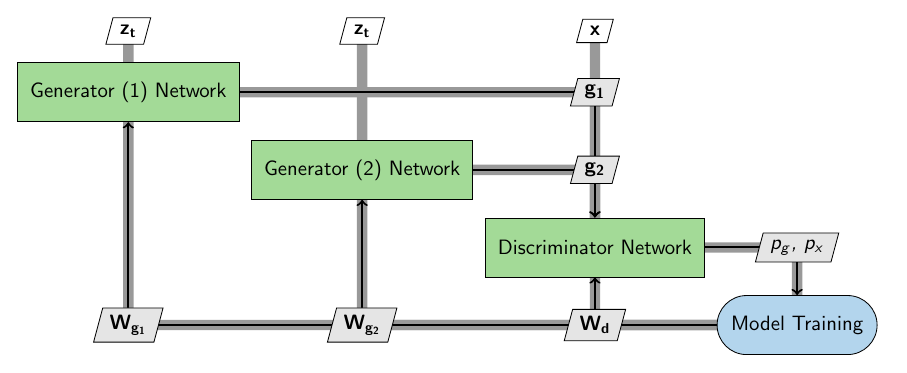}
    \vspace{-0.5cm}
    \caption{The twinGAN model serves as an intelligent parametrization method for realistic power and wing angle trajectories.
    The separate generators ensure the trajectories are generated independently.
    The inputs to the generators are the random noise variables ($\mathbf{z}_t$) .
    The reference data ($\mathbf{x}$) are passed to the discriminator along with the generated power ($\mathbf{g_1}$) and wing angle ($\mathbf{g}_2$) control points.
    The predicted probabilities ($p_g$ and $p_x$) are used by the optimizer to update the network weights ($\mathbf{W}_{g_1}$, $\mathbf{W}_{g_2}$, and $\mathbf{W}_d$).
    }
    \label{fig:TwinGAN_XDSM}
\end{figure}

The twinGAN~\cite{sisk2024surrogate} acts as an intelligent parametrization method, which generates the power and wing angle trajectories for the optimization.
The twinGAN reads random input variables and generates 40 B-spline control variables used to parameterize the takeoff trajectories (Fig.~\ref{fig:TwinGAN_XDSM}).
The twinGAN implicitly reduces the original design requirement space by generating only realistic trajectories and explicitly reduces the design requirement space down to three dimensions while maintaining sufficient variability as demonstrated in \citet{sisk2024surrogate} .
The separated generator networks of the twinGAN ensure the flexibility of the power and wing angle trajectory generation.
A minimax loss function is used during training as follows.
\begin{equation}
    \label{eq:TwinGAN}
    \min_{G} \max_{D} V\left(D,G\right)=E_{\mathbf{x} \sim p_{\text{data}}\left(\mathbf{x}\right)}\left[\log D\left(\mathbf{x}\right)\right]+E_{\mathbf{z} \sim p_\mathbf{z}\left(\mathbf{z}\right)}\left[\log \left(1-D\left(G_1\left(\mathbf{z}_t\right),G_2(\mathbf{z}_t)\right)\right)\right],
\end{equation}
where $V$ is the loss function, $G$ is the generator, $D$ is the discriminator, $\mathbf{x}$ is from the training data $p_{data}$, and $\mathbf{z}_t$ is from the random variable distribution $p_z$ (\emph{i.e.}, Uniform(0, 1) in this work).

As mentioned above, the twinGAN has two generators to generate power and wing angle trajectories separately for flexibility.
Each generator has four hidden layers with $[125,256,512,1024]$ nodes per layer respectively.
The generators each have three input noise variables and 20 output B-spline control points.
The twinGAN was trained on 1,099 optimal trajectories, which were gathered by varying the mass and propulsive efficiency of the simulation model~\cite{sisk2024surrogate}.
The mass is varied by $\pm 15\%$ from the baseline value of 725~kg, while the efficiency is varied $\pm10\%$ from the baseline of 90\%.

\subsection{Surrogates}
\label{sec:Surrogates}
\begin{figure}[htbp!]
    \centering
    \includegraphics[scale=0.2]{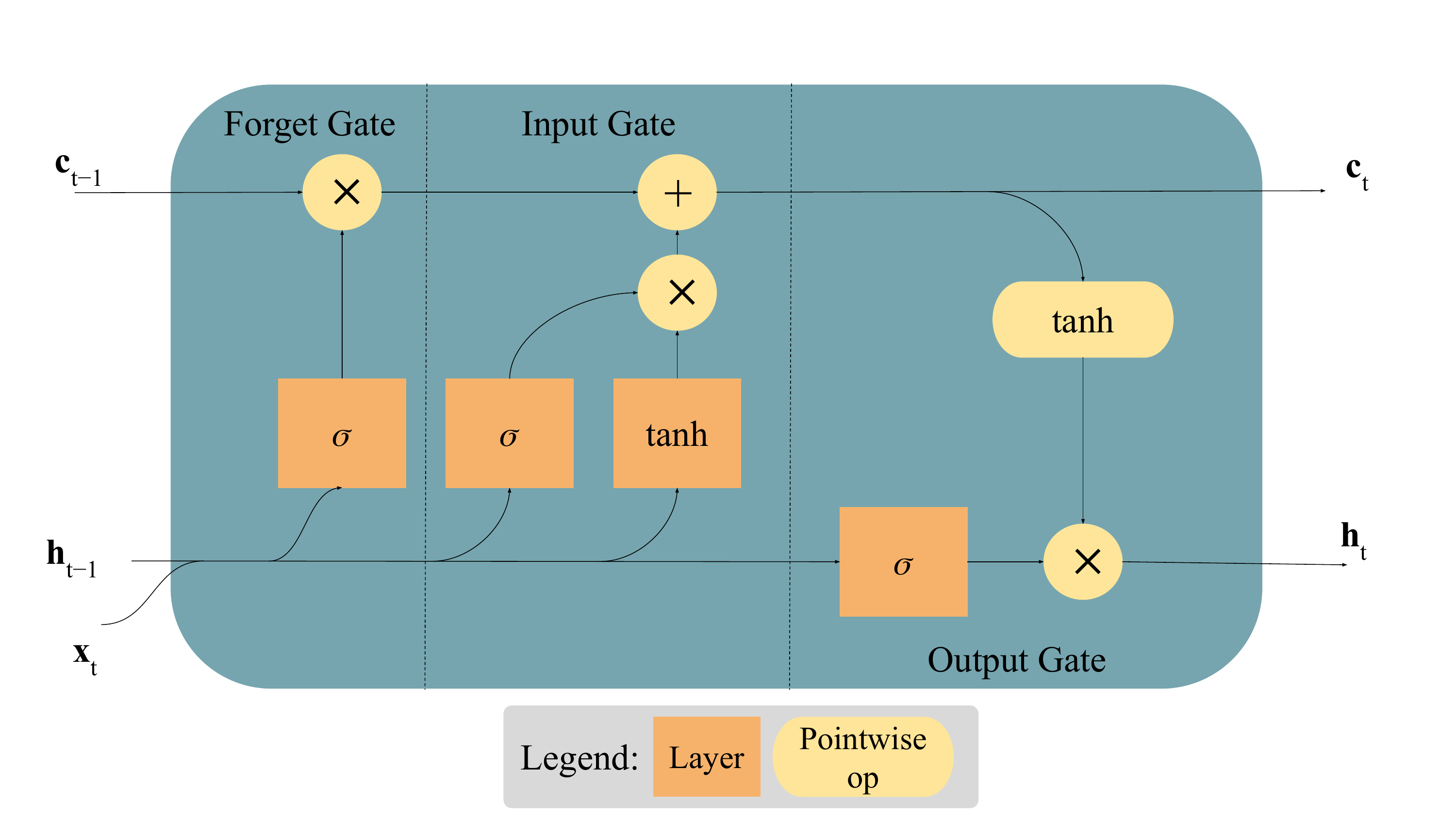}
    \caption{The LSTM cell is differentiated from a neural network node by the internal memory state of the cell, as well as the gates which modify the memory of the cell. These adaptations allow LSTM cells to retain information from long time steps back, which aids their ability to predict time-series data.}
    \label{fig:LSTMcell}
\end{figure}

This work uses long short-term memory (LSTM) networks for predicting time-series quantities of interest. 
An LSTM cell (Fig.~\ref{fig:LSTMcell}) contains an internal memory state, which is modified by three internal gates, \emph{i.e.}, the forget ($\mathbf{f}_t$), input ($\mathbf{i}_t$), and output ($\mathbf{o}_t$) gates.
The gates modify what is forgotten from, what is added to, and what is output from the cell memory, which aids time-series prediction~\cite{hochreiter1997long}.
The gates are made up of combinations of \texttt{tanh} and \texttt{sigmoid} functions, which, respectively, propose and select modifications to the internal memory state.
The equations for each gate are shown below, 
\begin{equation}
    \label{eq:InputGate}
    \mathbf{i}_t = \sigma \left(\mathbf{W}_i[\mathbf{h}_{t-1},\mathbf{x}_t]+b_i \right),
\end{equation}
\begin{equation}
    \label{eq:ForgetGate}
    \mathbf{f}_t = \sigma \left(\mathbf{W}_f[\mathbf{h}_{t-1},\mathbf{x}_t]+b_f \right),
\end{equation}
\begin{equation}
    \label{eq:OutputGate}
    \mathbf{o}_t = \sigma \left(\mathbf{W}_o[\mathbf{h}_{t-1},\mathbf{x}_t]+b_o \right),
\end{equation}
where $\sigma$ is the \texttt{sigmoid} function, $\mathbf{W}_x$ represents the weights for the given gate, $b_x$ represents the biases for the given gate, $\mathbf{h}_{t-1}$ is the cell output for the previous time step, and $\mathbf{x}_t$ is the input for the current time step.
The remainder of the cell operations are shown in the equations below,
\begin{equation}
    \label{eq:CellState}
    \Tilde{\mathbf{c}}_t = \tanh{(\mathbf{W}_c[\mathbf{h}_{t-1},\mathbf{x}_t]+b_c)},
\end{equation}
\begin{equation}
    \label{eq:CellOutput}
    \mathbf{h}_t = \mathbf{o}_t \odot \tanh{(\mathbf{c}_t)},
\end{equation}
\begin{equation}
    \label{eq:CandidateCellState}
    \mathbf{c}_t = \mathbf{f}_t \odot \mathbf{c}_{t-1} + \mathbf{i}_t \odot \Tilde{\mathbf{c}}_t,
\end{equation}
where $\Tilde{\mathbf{c}_t}$ is the candidate for the cell state, $\mathbf{c}_t$ is the current cell state, $\mathbf{h}_t$ is the cell output for the current time step, and $\odot$ is Hadamard product operation.

The surrogates predict the states ($E$, $\mathbf{x}$, $\mathbf{y}$, $\mathbf{v}_x$, $\mathbf{v}_y$, $\mathbf{a}$) from the simulation model for the duration of the takeoff process, where $E$ is the total energy consumed during takeoff, $\mathbf{x}$ and $\mathbf{y}$ are the displacements, $\mathbf{v}_x$ and $\mathbf{v}_y$ are the velocities, and $\mathbf{a}$ is the acceleration of the aircraft.
Specifically, a deep neural network (DNN) surrogate is used for predicting the scalar energy and has six input variables, one output variable, and five hidden layers featuring $[410,290,500,310,500]$ nodes on each layer respectively.
Each time-series model response is predicted by a LSTM network.
The LSTM models have six input variables, 500 output variables, and four hidden layers of 200 cells each.

\subsection{Physics-Constrained Generative Adversarial Network}
\label{sec:ConGAN}
\begin{figure}[htbp!]
    \centering
    \includegraphics[scale=0.7]{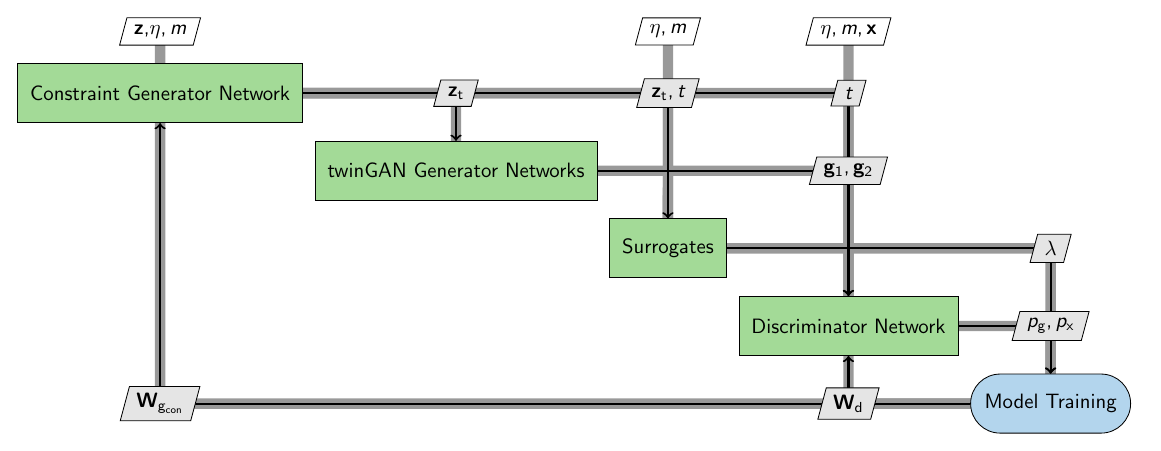}
    \caption{
    The inputs to the generators ($\mathbf{z}$, $\eta$, and $m$) are the random noise variables, efficiency, and mass.
    The predicted twinGAN noise variables ($\mathbf{z}_t$) are passed to the twinGAN generators to predict the generated power and wing angle control points.
    The reference data and labels ($\mathbf{x}$, $\eta$, and $m$) are passed to the discriminator along with the predicted flight time ($t$), generated power ($\mathbf{g}_1$), and wing angle ($\mathbf{g}_2$) control points.
    The penalty ($\lambda$) is based on the surrogate-predicted feasibility of the trajectory.
    The predicted probabilities ($p_g$ and $p_x$) are used by the optimizer to update the network weights ($\mathbf{W}_{g_{con}}$ and $\mathbf{W}_d$).
    }
    \label{fig:ConGAN_XDSM}
\end{figure}

The physicsGAN is developed as an intelligent parametrization method, which transforms the original design requirement space to a feasible design requirement space to facilitate the optimization process.
The architecture of the physicsGAN is shown in Fig~\ref{fig:ConGAN_XDSM}.
The physicsGAN also uses the twinGAN generator networks to generate power and wing angle trajectories separately for flexibility.
Training the physicsGAN model requires a feasible dataset where all trajectories are feasible. 
This training dataset is created using random samples generated by the twinGAN model and verified by surrogates for constraint feasibility. 
The physicsGAN uses design requirements, \emph{i.e.}, mass ($m$) and power efficiency ($\eta$) as labels, which are passed to the discriminator networks for adversarial training and the surrogate models for penalizing trajectories that violate constraint(s).
The twinGAN generator networks share the same architecture as twinGAN, using four layers with $[125,256,512,1024]$ nodes respectively.
Each layer has a \texttt{Leaky\_ReLU} activation function with $\alpha = 0.2$ and batch normalization with the momentum set to 0.8.
The constraint generator outputs takeoff duration ($t$) as well as three variables with a \texttt{sigmoid} activation function, which represent the noise variables for the twinGAN ($\mathbf{z}_t$).
The discriminator network is composed of three layers with $[1024,512,256]$ nodes respectively and outputs a single value.
The discriminator layers are followed by a \texttt{Leaky\_ReLU} activation function with $\alpha = 0.2$.
The output of the discriminator also passes through a \texttt{sigmoid} activation function.
Thus, the minimax loss function used for training is as follows.
\begin{equation}
    \label{eq:ConGAN}
    \min_{G} \max_{D} V\left(D,G\right)=E_{\mathbf{x} \sim p_{\text{data}}\left(\mathbf{x}\right)}\left[\log D\left(\mathbf{x}\right)\right]+E_{\mathbf{z} \sim p_\mathbf{z}\left(\mathbf{z}\right)}\left[\log \left(1-D\left(G_1\left(G_{con}\left(\mathbf{z}\right)\right),G_2\left(G_{con}\left(\mathbf{z}\right)\right)\right)\right)\right],
\end{equation}
where $V$ is the loss function, $G$ is the generator, $D$ is the discriminator, $\mathbf{x}$ is from the training data $P_{data}$, and $\mathbf{z}$ is from the random variable distribution $P_z$ (\emph{i.e.}, Uniform(0, 1) in this work).
Moreover, the penalty for the generator is based off the predicted violation of the acceleration constraint.
To calculate the penalty, the generated twinGAN noise variables, flight duration, efficiency, and mass are passed to the surrogate models for predicting design constraints. 
The constraint generator receives penalty if generated trajectories violate acceleration constraint using $\lambda = (a_{\text{max}} - 0.3)^2 \hspace{1mm}$ {if} $a_{\text{max}} \ge 0.3$; otherwise, no penalty needed. 

\section{Results}
\label{sec:results}
This section demonstrates the physicsGAN on the takeoff trajectory design of the $A^3$ Vahana drone and compares results with simulation-based optimal design as well as twinGAN performance.

\subsection{Problem Formulation}
The trajectory optimization aims to complete a successful vertical takeoff while minimizing the energy consumed (Table~\ref{tab:Optimization Problem Statement}).
The design variables for the optimization are B-spline control points for simulation-based design, the random noise variables for twinGAN, or random noise variables of physicsGAN.
The optimization problem is subject to constraints on the final displacement and velocity to ensure the aircraft reaches cruise conditions, as well as path constraints on the minimum vertical displacement for ground avoidance and maximum acceleration during takeoff for passenger comfort and safety.
The physicsGAN is trained to exploit the feasible design requirement space.
In this work the power efficiency ($\eta$) and mass ($m$) of the system are flight conditions to be varied.
\begin{table}[htbp!]
    \caption{Takeoff trajectory design problem formulation.}
    \label{tab:Optimization Problem Statement}
    \centering
    \begin{tabular}{c c c c}
        \hline
        & \textbf{Function/Variable} & \textbf{Description} & \textbf{Quantity} \\
        \hline
        minimize & \textit{E} & Electrical energy & \\
        with respect to & $\textbf{x}$, $\textbf{z}_t$, or $\textbf{z}$ & B-spline control points, twinGAN variables, or physicsGAN variables & 40, 3, or 3 \\
        & $t$ & Total takeoff time & 1 \\
        & & \textbf{Total design variables} & \textbf{41}, \textbf{4}, or \textbf{3} \\
        subject to & {$x_{f} \ge 900$ m} & Final horizontal displacement constraint & 1 \\
        & {$y_{f} \ge 305$ m} & Final vertical displacement constraint & 1 \\
        & {$v_{x,f} \ge 67 \frac{\text{m}}{\text{s}}$} & Final horizontal speed constraint & 1 \\
        & {$y_{\text{min}} \ge 0$ m} & Vertical displacement constraint & 1 \\
        & {$a_{\text{max}} \le 0.3$ g} & Acceleration constraint & 1 \\
        & & \textbf{Total constraint functions} & \textbf{5} \\
        \hline
    \end{tabular}
\end{table}

\subsection{Surrogate Modeling}
The twinGAN-based and physicsGAN-based optimizations both use surrogate models to predict the state variables during takeoff.
Specifically, the surrogates use the twinGAN noise variables ($\mathbf{z}_t$) as well as the labels ($\eta$ and $m$) and takeoff duration (${t}$) as input.
The takeoff duration used to train the surrogates is bounded within [25, 35] seconds, which is representative of the optimal trajectories used to train the twinGAN.
The data for the surrogates comes from multi-disciplinary analysis. 
They are trained on 5,504 samples, validated on 1,376 samples, and tested on 2,573 samples.
The surrogate models achieve over 99\% accuracy for almost all the quantities of interest (Table~\ref{tab:Sum_surr_acc}).
\begin{table}[htbp!]
    \caption{Summary of surrogate model testing accuracy.}
    \label{tab:Sum_surr_acc}
    \centering
    \begin{tabular}{c c c c c c}
        \hline
        Model Response & $E$ & $\mathbf{x}$ & $\mathbf{y}$ & $\mathbf{v}_x$ & $a$ \\
        \hline
        Accuracy & 99.90 \% & 98.78\% & 99.46 \% & 99.27\% & 99.87 \% \\
        \hline
    \end{tabular}
\end{table}

\subsection{physicsGAN Verification}
The physicsGAN is trained following the process in Fig.~\ref{fig:ConGAN_XDSM}.
The generator and discriminator of the physicsGAN are trained with a batch size of 107 and utilize 10,601 feasible training samples.
The performance of physicsGAN is assessed using the coverage percentage, which represents the percentage of constraint satisfaction in the transformed feasible design space.
It is calculated by randomly sampling the design requirements (\emph{i.e.}, $\eta$ and $m$) and design variables (\emph{i.e.}, physicsGAN noise variables) and checking the percentage of samples that meet the constraints.
To ensure adequate sampling, $10000*N$ samples are gathered where $N$ is the number of noise variables for each physicsGAN model. 
The physicsGAN model with the highest coverage percentage during training is saved for surrogate-based optimization.

Additionally, to ensure adequate trajectory variants are generated by the physicsGAN, a fitting optimization is conducted using the \texttt{Powell}~\cite{Powell1964efficient} optimizer in \texttt{SciPy}~\cite{SciPy}.
Specifically, fitting accuracy is computed as follows based on the agreement between existing, training trajectories and the reconstructed counterparts by physicsGAN (Table~\ref{tab:fitting}).
\begin{equation}
    \label{eq:14}
    RelAcc = \left(1 - \frac{1}{N*N_c} \sum_{{i}=1}^{N} \frac{\norm{\mathbf{g}_{\text{pred}_{i}} - \mathbf{g}_{\text{true}_{i}}}_1}{\norm{\mathbf{g}_{\text{true}_{i}}}_1}\right)100\%,
\end{equation}
where $N$ = 10,601 is the total number of training trajectories, each of which contains $N_c$ = 40 B-spline control points originally used for parameterizing takeoff trajectories, $\mathbf{g}_{\text{pred}_{i}}$ are the predicted control points, and $\mathbf{g}_{\text{true}_{i}}$ are the reference control points.
The physicsGAN with three noise variables achieves more than 99.5\% fitting accuracy and close to 99\% coverage while twinGAN exhibits less than 5\% coverage (Table~\ref{tab:fitting}).
Moreover, physicsGAN consistently achieves around 99\% fitting accuracy and more than 99\% coverage with one or two noise variables. 
\begin{table}[htbp!]
    \centering
    \caption{Comparison of twinGAN and physicsGAN fitting accuracies and coverage percentage.}
    \begin{tabular}{c c c c c}
        \hline
        Model & Noise Variables & Power Fitting Accuracy & Wing Angle Fitting Accuracy & Coverage Percentage \\
        \hline
        twinGAN & 3 & 99.18\% & 98.99\% & 4.64\% \\
        \hline
        \multirow{3}{*}{physicsGAN} & 1 & 98.78\% & 99.43\% & 99.79\% \\
                                   & 2 & 99.51\% & 99.61\% & 99.20\% \\
                                   & 3 & 99.59\% & 99.64\% & 98.85\% \\
        \hline
    \end{tabular}
    \label{tab:fitting}
\end{table}

To demonstrate the performance, physicsGAN-based takeoff trajectory design is investigated at the center of the design requirement space ($\eta = 0.893 \hspace{0.2cm}\text{and}\hspace{0.2cm} m=723.85$ kg).
The physicsGAN leverages the \texttt{Differential Evolution} optimizer in \texttt{SciPy} and manages to match the simulation-based reference optimal design (Table~\ref{tab:ConGAN_parametric_study}).
The twinGAN framework is tested with the \texttt{Differential Evolution} optimizer and the gradient-based \texttt{Trust-Constr} optimizer in \texttt{SciPy}.
With the \texttt{Differential Evolution} optimizer the twinGAN-based optimal design matches the simulation-based optimal design (reference) with an accuracy of 95.19\%, and the twinGAN-based optimization takes 21.93 seconds.
When varying the random seed with the \texttt{Trust-Constr} optimizer, results present that the twinGAN-based optimal design matches the simulation-based optimal design (reference) with an average accuracy of 85.54\%, and the twinGAN-based optimization takes 25.67 seconds on average.
However, the twinGAN-based optimal designs shown in Table~\ref{tab:ConGAN_parametric_study} have been violating the acceleration constraint.
For these cases, the \texttt{Trust-Constr} optimizer exits the trust region, which triggers the optimizer's exit condition.
\begin{table}[!h]
    \caption{Parametric study of physicsGAN-based surrogate optimization accuracy and time.}
    \label{tab:ConGAN_parametric_study}
    \centering
    \begin{tabular}{p{2cm} p{2cm} p{1.5cm} p{2cm} p{1.5cm} p{1.5cm} p{1.5cm}}
        \hline
        Optimization Framework & Optimizer & Design Variables & Energy (Wh) 
        & Relative Accuracy & Constraint Violation & Time (s) \\
        \hline
        Reference & \texttt{SNOPT} & 41 & 1883.84 & N/A & Satisfied & 533.38 \\
        \hline
        \multirow{4}{2cm}{twinGAN} & \multirow{4}{2cm}{\texttt{Trust-Constr}} & \multirow{4}{1.5cm}{4} 
        & 1693.54 & 89.90 \% & 31.96 \% & 30.39 \\                                                         
        & & & 1343.20 & 71.30 \% & 23.51 \% & 28.59 \\
        & & & 1812.25 & 96.20 \% & 0.05 \% & 6.69 \\
        & & & 1597.11 & 84.78 \% & 54.04 \% & 37.00 \\
        \hline
        twinGAN & \texttt{Differential Evolution} & 4 & 1793.29 & 95.19 \% & 4.70 \% & 21.93 \\
        \hline
        \multirow{3}{2cm}{physicsGAN} & \multirow{3}{2cm}{\texttt{Differential Evolution}} 
        & 1 & 1986.16 & 94.57 \% & Satisfied & 0.98 \\
        & & 2 & 1947.47 & 96.62 \% & Satisfied & 2.95 \\
        & & 3 & 1891.02 & 99.62 \% & Satisfied & 2.25 \\
        \hline
    \end{tabular}
\end{table}

\begin{figure}[!h]
    \centering
    \adjustbox{trim=0.7cm 0.1cm 0.6cm 0.5cm,clip=true}{ 
    \includegraphics[scale=0.7]{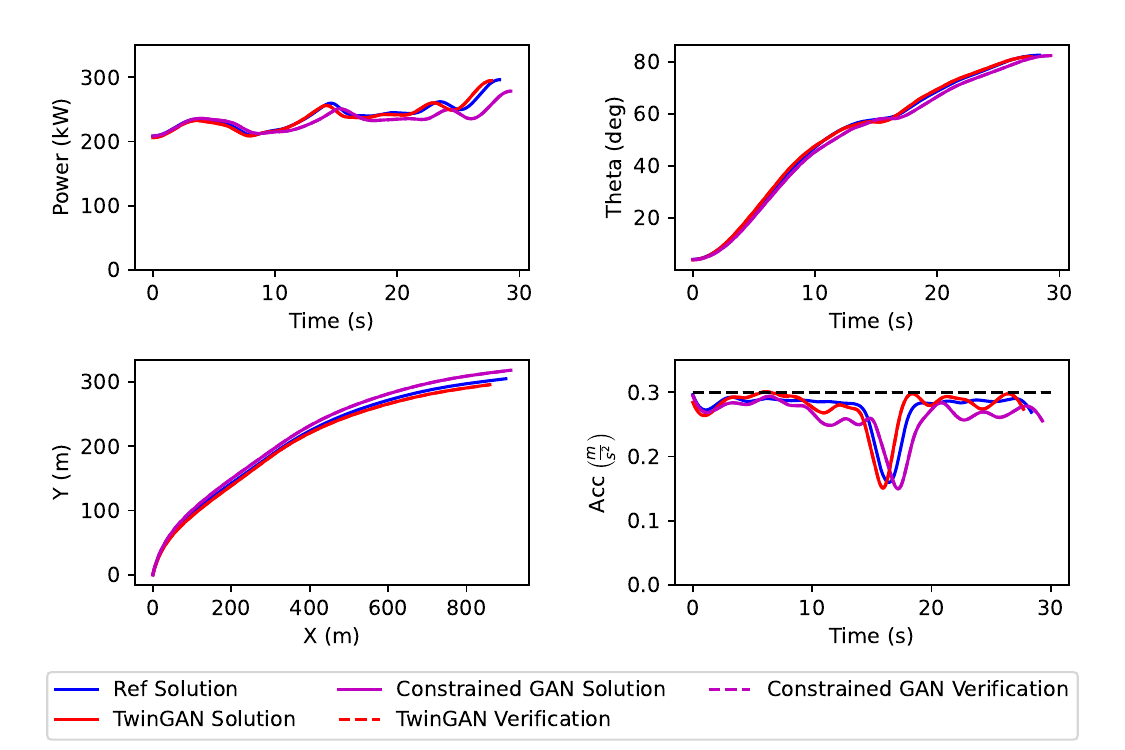}}
    \caption{The proposed framework generates control curves power (top left) and wing angle (top right) while maintaining feasibility. 
    These results correspond to the best seed for the twinGAN from Table~\ref{tab:ConGAN_parametric_study}.}
    \label{fig:traj}
\end{figure}

\begin{figure}[!h]
    \centering
    \adjustbox{trim=0.7cm 0.1cm 0.6cm 0.5cm,clip=true}{ 
    \includegraphics[scale=0.7]{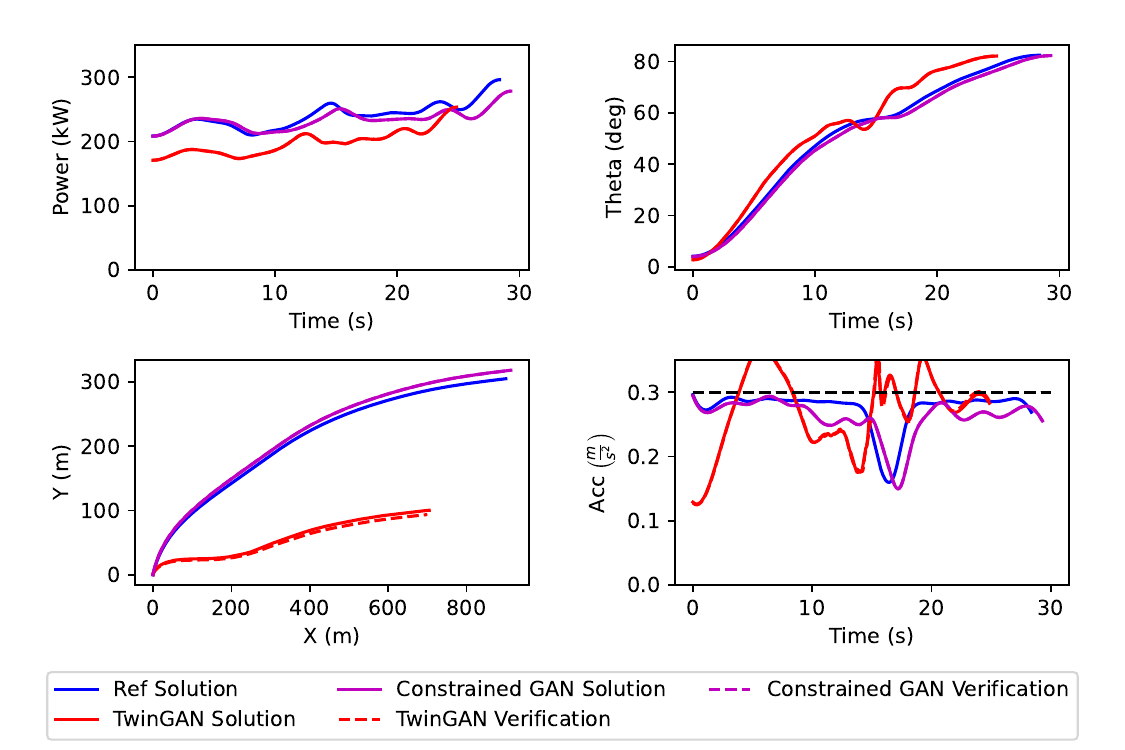}}
    \vspace{-0.2cm}
    \caption{These results demonstrate how a bad initial seed for the twinGAN produces an infeasible and inaccurate trajectory, which violates the acceleration constraint by over 20\%.}
    \label{fig:traj2}
\end{figure}

The best physicsGAN results occur with three noise variables.
The physicsGAN-based optimization matches the simulation-based optimal design (reference) with 99.62\% accuracy and maintains the feasibility even without using explicit constraints for optimization.
Additionally, the physicsGAN-based optimization is 200 times faster than the simulation-based optimal design (reference).
Moreover, the physicsGAN-based optimization outperforms twinGAN-based optimization in terms of efficiency while in the meantime twinGAN-based optimization failed to identify the feasible, optimal design. 
The reduction in computational cost by the physicsGAN is due to the reduced design requirement space dimension, as well as the removal of the constraints.
The computational cost of evaluating the constraints is problem and optimizer dependent, but the physicsGAN fully removes this cost.
Visualization cases for comparison are shown in Figs.~\ref{fig:traj} and \ref{fig:traj2}.

The best trajectories for each framework, shown in Fig.~\ref{fig:traj}, show close alignment with all three takeoff trajectories sharing similar shapes.
The optimizer tends to minimize the takeoff duration, since the energy consumed is the integral of the power with respect to time.
This results in a steady increase in the power applied to the motors and the angle of the wing, as well as the acceleration being maximized.
However, all three models show a consistent reduction in the acceleration around the 17-18 second mark.
This dip shows the optimizer making a trade-off between a faster trajectory and a trajectory with lower power to minimize the energy.
Figure~\ref{fig:traj2} shows a set of results where twinGAN-based optimization gets stuck in infeasible region.
For this case, the twinGAN-based optimization stopped with infeasible results and exited due to leaving the trust region.
This shows that the small feasible region produced by the twinGAN causes the framework to lack robustness.

\section{Conclusions}
\label{sec:conclusion}
Urban air mobility (UAM) is growing rapidly as the future transportation of humans.
Electric vertical takeoff and landing (eVTOL) aircraft are expected to serve as the primary method of transportation for UAM.
Thus, eVTOL aircraft aim to provide hub to hub transportation in urban areas, which will alleviate the congestion urban cities face.
However, eVTOL adoption has a few barriers, among which the limited range and endurance due to battery density is critical for practical applications.
Multidisciplinary analysis and optimization (MDAO) can determine efficient takeoff trajectories by balancing the coupled disciplines in the analysis but often incurs a large computational cost due to iteratively evaluating the simulation model.

This work proposed a novel, physics-constrained generative adversarial network (physicsGAN), which transformed the original design requirement space into a feasible, reduced-dimension design requirement space.
As far as the literature review shows, this is the first work incorporating physics constraints in a generative artificial intelligence framework.
Using only three noise variables the proposed framework achieved a feasible space, which greatly alleviated the difficulty of surrogate-based optimization and accelerated the trajectory optimization compared to the simulation-based reference method.
Operating on the feasible space enables rapid, robust, accurate trajectory optimization.

Moreover, the proposed physicsGAN were also compared with data-driven generative adversarial networks (GAN). 
The results revealed that physicsGAN achieved higher accuracy and efficiency while the data-driven GAN-based design optimization could fail due to getting stuck in an infeasible region. 
In real practice, including eVTOL aircraft and self-driving cars, an infeasible design can cause disastrous accidents.

The proposed physicsGAN-enabled surrogate-based optimization in this work was only tested on the acceleration constraint.
A physicsGAN with more constraints will be investigated in the future work to provide a better comparison.

\bibliography{phlai,sample}

\end{document}